\documentclass[letterpaper]{article} 
\usepackage[preprint]{aaai2027}  
\usepackage[hyphens]{url}  
\usepackage{graphicx} 
\usepackage{natbib}  
\usepackage{caption} 
\usepackage{algorithm}
\usepackage{algorithmic}
\usepackage{dblfloatfix}
\usepackage{amsmath}
\usepackage{newfloat}
\usepackage{listings}
\DeclareCaptionStyle{ruled}{labelfont=normalfont,labelsep=colon,strut=off} 
\floatstyle{ruled}
\newfloat{listing}{tb}{lst}{}
\floatname{listing}{Listing}

\usepackage{booktabs}
\usepackage{amsfonts} %
\title{Beyond Task-Only Matching: Personalized Skill Routing with Counterfactual Evaluation}
\author{
    Written by AAAI Press Staff\textsuperscript{\rm 1}\thanks{With help from the AAAI Publications Committee.}\\
    AAAI Style Contributions by Peter Patel Schneider,
    Sunil Issar,\\
    J. Scott Penberthy,
    George Ferguson,
    Hans Guesgen,
    Francisco Cruz\equalcontrib\corresponding,
    Marc Pujol-Gonzalez\equalcontrib\corresponding
}
\affiliations{
    \textsuperscript{\rm 1}Association for the Advancement of Artificial Intelligence\\

    1101 Pennsylvania Ave, NW Suite 300\\
    Washington, DC 20004 USA\\
    proceedings-questions@aaai.org
}

\title{Beyond Task-Only Matching: Personalized Skill Routing with Counterfactual Evaluation}
\author {
    Tianle Wang\textsuperscript{\rm 2}\equalcontrib,
    Yanghe Zou\textsuperscript{\rm 1}\equalcontrib,
    Xiang Liu\textsuperscript{\rm 1,\rm 3},
    Ziyao Huang \textsuperscript{\rm 4},
    Chenchen Fu \textsuperscript{\rm 1},
    Weiwei Wu \textsuperscript{\rm 1}
}
\affiliations {
    \textsuperscript{\rm 1}School of Computer Science and Engineering, Southeast University \\
    \textsuperscript{\rm 2}College of Software Engineering, Southeast University\\
    \textsuperscript{\rm 3}Department of Computer Science and Engineering, The Chinese University of Hong Kong\\
    \textsuperscript{\rm 4}Department of Computer Science, City University of Hong Kong\\
    
}

\begin{document}

\maketitle

\begin{abstract}
The rapid expansion of reusable skill repositories makes skill routing a critical capability for large language model (LLM) agents. Existing methods treat routing as task-only semantic matching.
However, when users with incompatible constraints issue an identical request, this assumption conflates task relevance with skill suitability: a task-only router can select a semantically plausible skill that is unsuitable for the requesting user. 
To expose this failure mode, we formulate \textit{personalized skill routing} as profile-conditioned retrieval, in which relevance depends jointly on the task and the user profile. We first introduce a profile-counterfactual benchmark, in which the task is held fixed while changes in the user profile induce changes in the reference skill. We further construct paired counterfactual supervision and propose SkillFeed, a progressive retrieve-and-rerank framework that first establishes task--skill alignment and then learns profile-conditioned discrimination. By retrieving body-level evidence and reranking semantically similar but profile-conflicting candidates, SkillFeed identifies skills that satisfy both task requirements and user constraints. 
On SkillFeed-Bench, SkillFeed attains 75.1\% top-1 retrieval accuracy, a 23.1-point improvement over the corresponding pretrained routing baseline. Adding profile conditioning yields a 35.1-point gain on queries where user profile changes the reference skill. This contrast shows that user profiles are most consequential precisely when they change skill suitability. Our website is publicly available at \url{http://www.aiskillfeed.com}.
\end{abstract}

\section{Introduction}
\label{sec:intro}
Large language model (LLM) agents increasingly reason, call tools, and execute multi-step workflows~\citep{yao2023react,schick2024toolformer,shen2023hugginggpt}. In these systems, \emph{skills} package reusable procedures, tool-use strategies, and execution constraints in versioned bundles with \texttt{SKILL.md} manifests~\citep{openai2026skills,zheng2026skillrouter}. As public repositories grow, an agent must route each request to the skill that best supports its intended execution, making skill routing a central retrieval problem~\citep{li2025skillflow,zheng2026skillrouter}.

\begin{figure}[t]
    \centering
    \includegraphics[width=0.84\columnwidth]{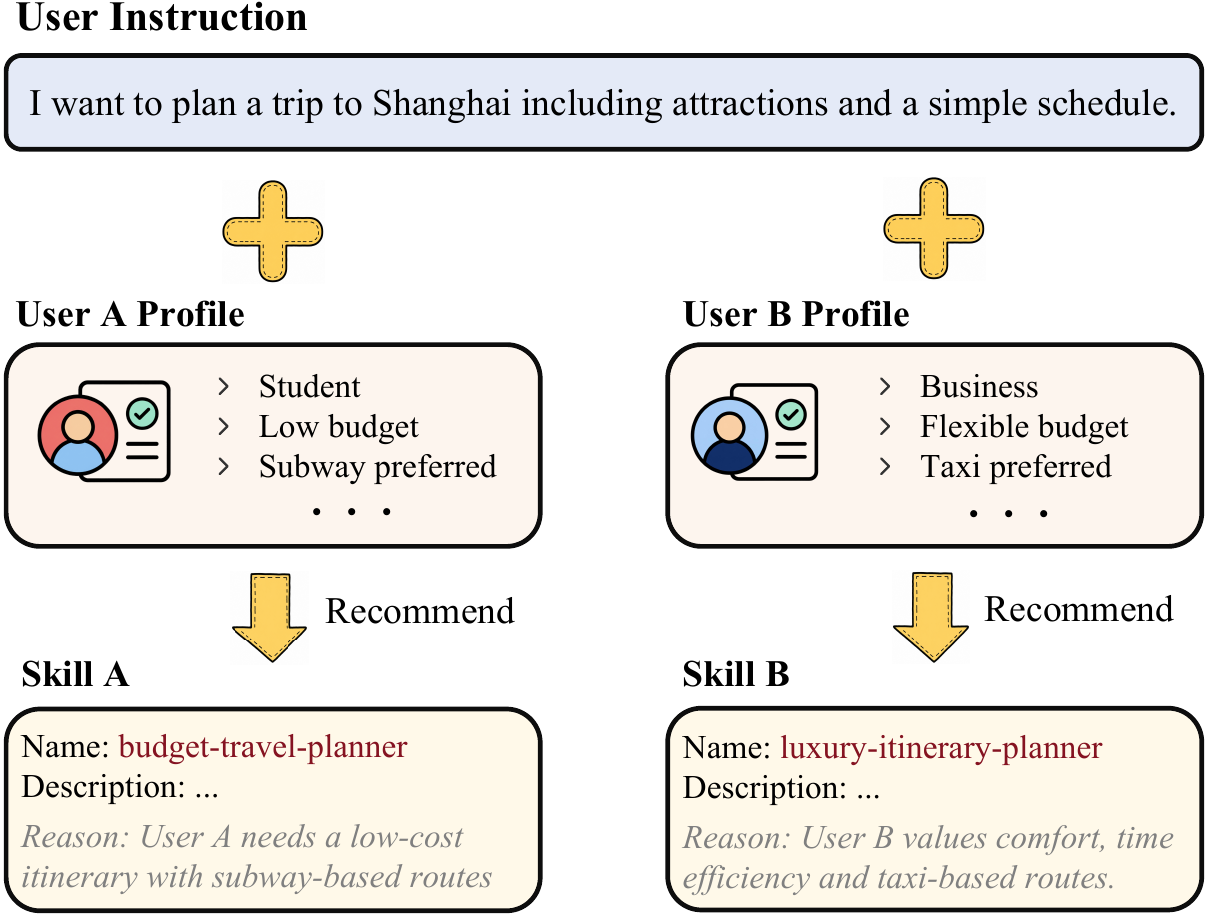}
    \caption[Example of personalized skill routing.]{Motivation example of personalized skill recommendation.
    The same task, "planning a trip to Shanghai," is routed to a budget travel planner for a student and to a luxury itinerary planner for a business traveler.
    The target skill changes with user profile.}
    \label{fig:motivation}
\end{figure}

Most existing methods cast skill routing as task-to-skill semantic matching~\citep{zheng2026skillrouter,li2025skillflow,wang2026r3skill}. This formulation assumes that the task uniquely determines the appropriate skill. In practice, however, a task describes \emph{what} a user wants to accomplish, whereas a user profile can determine \emph{how} that task should be carried out. When two users issue the same request under incompatible constraints, a task-only router may select a skill that is semantically relevant but unsuitable for one of them. For example, a workflow that relies on an unavailable platform, exceeds a budget, or assumes expertise the user does not have. As illustrated in Figure~\ref{fig:motivation}, the same travel-planning request calls for a budget itinerary skill for a student and a premium itinerary skill for a business traveler. The task is unchanged, but the appropriate skill changes with profile attributes such as budget, payment preference, platform ecosystem, and professional background.

This failure mode cannot be established with conventional task-level evaluation. Prior benchmarks for skill learning, execution, and routing pair a request with suitable skills, while existing retrieve-and-rerank systems are trained and evaluated primarily on task--skill relevance~\citep{li2026skillsbenchbenchmarkingagentskills,zheng2026skillrouter,wang2026r3skill}. Such evaluation cannot distinguish a system that truly uses a profile from one that ignores it: when the task remains fixed, it never asks whether the predicted skill should change as the user profile changes. The resulting data also provide limited supervision for separating skills that are equally relevant to the task but conflict in their compatibility with user constraints.

We therefore formulate \emph{personalized skill routing} as profile-conditioned retrieval, in which skill relevance is jointly determined by a task and its task-relevant user profile. This setting raises three coupled requirements: (1) a benchmark must isolate profile-induced changes in the reference skill; (2) training data must teach the model to discriminate among profile-conflicting alternatives; (3) and the routing system must scale to large repositories while using the fine-grained applicability conditions encoded in skill bodies. 
We meet these requirements by constructing controlled profile-counterfactual instances that hold the task fixed while varying the user profile and reference skill. These instances both evaluate whether routing responds to profile shifts and provide supervision for profile-conditioned retrieval and reranking.

Building on this formulation, we introduce SkillFeed-Bench and SkillFeed retrieval framework. SkillFeed-Bench evaluates profile-conditioned routing over a 228K-skill repository. SkillFeed first establishes task--skill alignment and then learns profile-conditioned discrimination; at inference time, it combines broad candidate recall with test-level evidence and profile-aware reranking. This design separates the need to preserve candidate coverage at repository scale from the need to resolve profile-specific suitability among semantically similar skills. Our contributions are as follows:

\begin{itemize}
  \item We formulate personalized skill routing as retrieval conditioned jointly on a task and a user profile. We introduce SkillFeed-Bench, a benchmark over 228K candidate skills with 329 annotated instances, including counterfactual profile-shift cases in which an unchanged task has different reference skills under different user profiles.
  
  \item We construct task-centric and counterfactual profile-conditioned training instances with semantically similar, profile-conflicting hard negatives. This supervision progressively adapts a dense retriever and trains a listwise reranker to recognize when a profile changes the suitable skill.

  \item We develop SkillFeed, a two-stage framework that combines dense and lexical candidate retrieval, profile-conditioned retrieval over skill chunks, reciprocal-rank fusion, and profile-aware reranking. 
  Experiments on SkillFeed-Bench show that SkillFeed achieves 75.1\% Hit@1,
    outperforming the pretrained baseline by 23.1\%. Profile conditioning contributes a 18.6-point gain on profile-sensitive queries, and 35.1-points on profile-counterfactual samples.
  
\end{itemize}

\section{Related Work}
\label{sec:related_work}

\paragraph{Personalized Retrieval and Recommendation.}
Personalized ranking models relevance as user-dependent, learning from
interaction histories~\citep{koren2009matrix,rendle2009bpr}, user profiles
as conditioning input for LLM-based recommenders~\citep{deldjoo2024reviewmodernrecommendersystems,
bang2026llmbaseduserprofilemanagement}. Personalized search and LLM agents further integrate
user memory and intent into ranking and long-term interaction~\citep{zhou2024cognitivepersonalizedsearchintegrating,
enablingpersonalizedlongterminteractions}. These methods establish the importance of conditioning ranking on user information, but their target is an item, document, or content preference rather than an executable procedure selected for a concrete task.In skill routing, a profile contains task-relevant constraints, such as language, platform access, budget, and expertise, that can make one workflow suitable and another inapplicable. Our focus is therefore not general personalization, but the profile-induced change in the suitable skill under a fixed task, which requires supervision and evaluation instances in which
changing the profile can change the reference target.

\paragraph{Skill Benchmarks.}
Recent benchmarks evaluate complementary aspects of agent skills, including skill-enabled execution~\citep{li2026skillsbenchbenchmarkingagentskills,han2026sweskillsbenchagentskillsactually}, lifelong skill discovery~\citep{zhang2026skillflowbench}, continual learning and skill generation~\citep{zhong2026skilllearnbench,zhou2026skillgenbench}, and large-scale skill retrieval~\citep{cho2026skillretlargescalebenchmarkskill,wang2026r3skill}. These resources provide valuable task-level supervision and evaluation, but they do not explicitly test whether a router should revise its selected skill when an unchanged task is paired with a different user profile. In contrast, SkillFeed-Bench includes controlled profile-shift instances in which the task is fixed while the annotated reference skill changes. This design separates profile sensitivity from ordinary task--skill semantic matching.

\paragraph{Skill Retrieval and Routing.}
Skill routing builds on sparse and dense retrieval methods, including BM25 and embedding-based retrieval~\citep{robertson2009bm25,wang2022e5,chen2024bge}. At repository scale, SkillRouter uses a retrieve-and-rerank pipeline, while R3-Skill studies query-conditioned routing with a two-stage retriever~\citep{zheng2026skillrouter,wang2026r3skill}; related work also explores retrieval augmentation for agentic AI~\citep{su2026skillretrievalaugmentationagentic}. These systems motivate efficient candidate generation and fine-grained reranking, but task relevance alone cannot distinguish skills that are semantically similar yet incompatible with different user constraints. Moreover, skill applicability may be expressed in the procedural body of a \texttt{SKILL.md} file through usage conditions, dependencies, and platform assumptions~\citep{openai2026skills,zheng2026skillrouter}. SkillFeed extends the retrieve-and-rerank paradigm by conditioning both candidate retrieval and final discrimination on the task-relevant user profile, using body-level evidence to resolve profile-conflicting alternatives.

\section{Problem Formulation and Benchmark}



\begin{figure*}[t]
    \centering
    \includegraphics[width=0.8\textwidth]{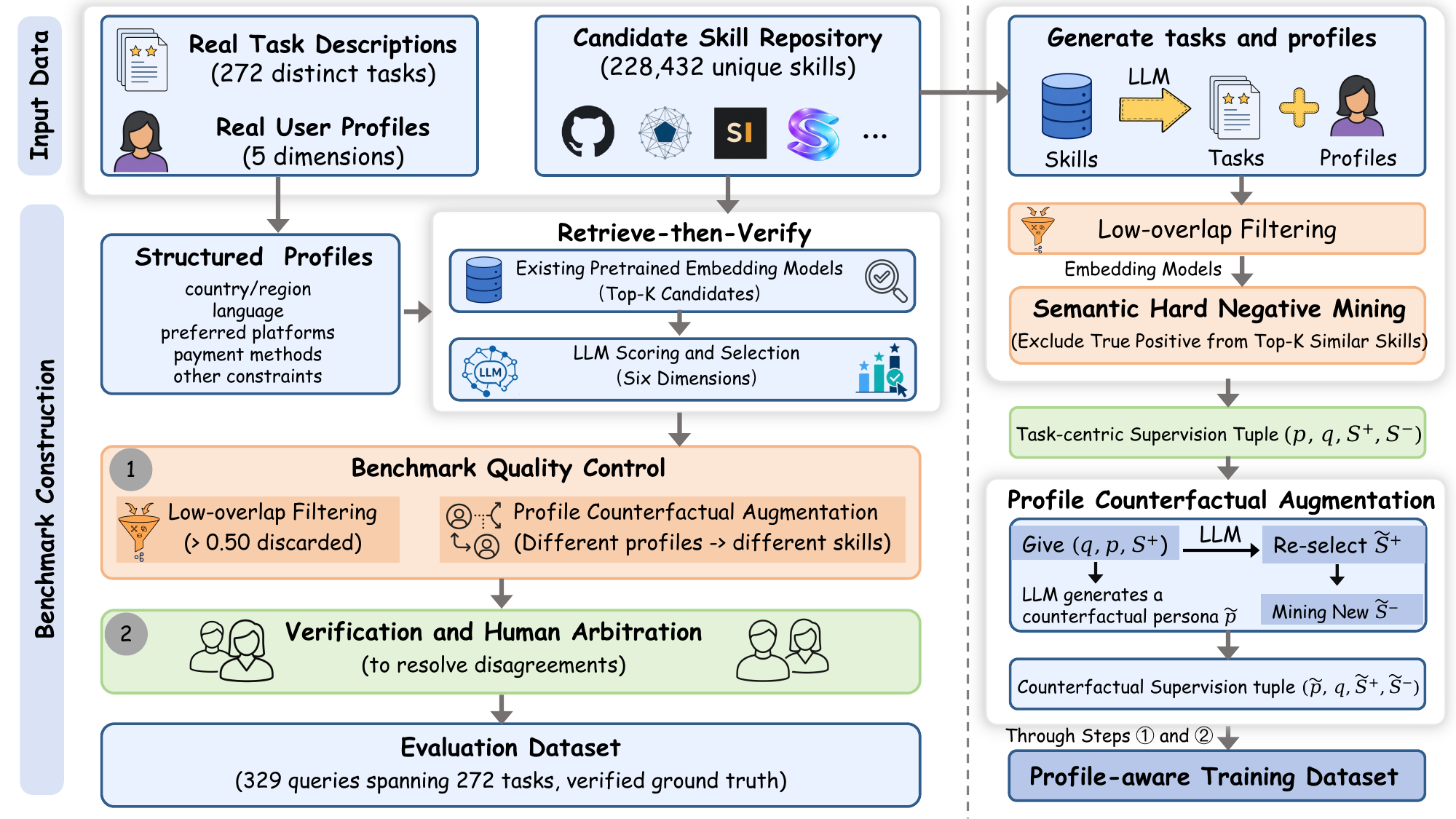}
    \caption{Overview of the dataset construction pipeline. }
    \label{fig:benchmark}
\end{figure*}
This section formalizes \textit{personalized skill routing}, a setting in which the correct skill is determined jointly by the task and the user profile rather than by task semantics alone. 
Under a fixed task, changing user profile constraints may change the annotated reference skill, which makes conventional task-only evaluation insufficient. 
Therefore, we instantiate this setting with SkillFeed-Bench, a benchmark built over 228,432 community-sourced skills and 329 evaluation queries spanning ten domains, each paired with a structured user profile and an annotated reference skill.




\subsection{Personalized Skill Routing Formulation}

Let $q$ denote a natural-language task query, $p$ a structured user profile, and $\mathcal{S}=\{S_1,\ldots,S_N\}$ a repository of candidate skills. The profile contains only task-relevant contextual attributes, such as country or region, language, platform access, payment method, expertise, and other constraints. Each candidate skill $S_i=(n_i,d_i,b_i)$ comprises a name $n_i$, a concise description $d_i$, and a procedural body $b_i$ that specifies its workflow and applicability conditions.

Personalized skill routing selects the candidate that best satisfies the task and profile jointly:
\begin{equation}
S_{q,p}^{*}
=
\arg\max_{S_i\in\mathcal{S}}
f(q,p,S_i),
\label{eq:personalized_skill_routing}
\end{equation}
where $f(q,p,S_i)$ measures the compatibility of $S_i$ with both $q$ and $p$. This differs from task-only routing: a skill can be semantically relevant to $q$ yet unsuitable because it conflicts with the profile's constraints.

For scalable retrieval, the system ranks all candidates by this score and returns the top-$K$ set:
\begin{equation}
\mathcal{R}_K(q,p)
=
\operatorname{TopK}_{S_i\in\mathcal{S}}
f(q,p,S_i),
\label{eq:personalized_topk_routing}
\end{equation}
where $\mathcal{R}_K(q,p)$ is the ordered set of the $K$ highest-ranked skills for $(q,p)$. Because each evaluation instance has one annotated reference skill, candidate recall succeeds when $S^*_{q,p}\in\mathcal{R}_K(q,p)$.

\subsection{SkillFeed-Benchmark Construction}

\paragraph{Data Sources.}
We aggregate skills from public sources, including GitHub, SkillNet~\cite{liang2026skillnetcreateevaluateconnect}, and SkillSMP~\cite{skillsmp2026}. After deduplication and format normalization, the repository contains 228,432 skills, each with a name, description, and procedural body.

\paragraph{Task and Profile Collection.}
We collect task descriptions from real LLM-agent interactions in which users sought help with specific goals. After manual screening for self-containedness, sufficient detail, and practical relevance, we retain 272 unique tasks. Each task is paired with a structured user profile covering country or region, language, preferred platforms, payment methods, and additional constraints such as regulatory requirements, budget limits, or domain-specific workflows. The profiles are designed to be concise while retaining attributes that can affect skill suitability.


\paragraph{Reference Target Annotation.}
\label{para:gt annotation}
Exhaustive annotation over 228,432 skills is impractical. We therefore use a two-stage annotation procedure. First, pretrained embedding models retrieve a top-$K$ set of semantically relevant candidates for each task--profile pair. Second, an LLM assesses these candidates using task relevance, profile compatibility, capability match, constraint satisfaction, specificity, and overfit risk, and then performs pairwise comparisons to select the reference skill. Each annotated instance is a triplet $(p,q,S)$ consisting of a profile, task query, and selected skill.


\paragraph{User Profile Counterfactual Augmentation.}

For an ordinary triplet $(p,q,S)$, the effects of the task and profile are coupled. We therefore construct counterfactual instances that hold $q$ fixed while varying $p$. Given an original instance, we prompt an LLM to generate a plausible alternative profile $\tilde{p}$ that deviates from the original $p$ and reapply the reference-target annotation procedure to $(\tilde{p},q)$. We retain the resulting triplet $(\tilde{p},q,\tilde{S})$ only when $\tilde{S}\neq S$. These retained pairs directly test whether a routing method changes its prediction when a profile change alters the suitable skill.

\paragraph{Quality Control.}

We apply quality controls to ensure that SkillFeed-Bench tests profile-conditioned routing rather than superficial task level semantic matching. We discard samples with confidence scores below 0.80 and task--skill pairs with lexical overlap above 0.50, reducing trivial keyword matches and evident annotation errors. We then apply consistency checks and manually review retained instances. At least two professional human annotators with prior experience in evaluating AI agents, software tools, or programming-related tasks independently evaluate every retained original and counterfactual instance; expert arbitration resolves disagreements, and persistently ambiguous cases are excluded. These procedures reduce the risk that benchmark performance reflects spurious lexical cues or annotation artifacts.

The resulting benchmark contains 329 evaluation queries over 228K candidate skills, including 162 profile-sensitive queries (77 profile-counterfactual augmentation) and 167 profile-insensitive queries. Sensitive and Insensitive subsets are defined by
presence of user context constraints.
Its controlled profile shifts expose a routing problem that task-only methods cannot resolve: a system must revise its selected skill when the user profile changes the suitable execution path, while still retrieving efficiently. 
SkillFeed-Bench is therefore architecture-agnostic, but it demands balancing  broad recall for its vast pool with fine-grained discrimination for user profile shifts. We next present SkillFeed, a two-stage retrieval-and-reranking approach designed to meet these requirements and evaluated on this benchmark.

\section{Method: SkillFeed}

\begin{figure}[t]
    \centering
    \includegraphics[width=0.98\columnwidth]{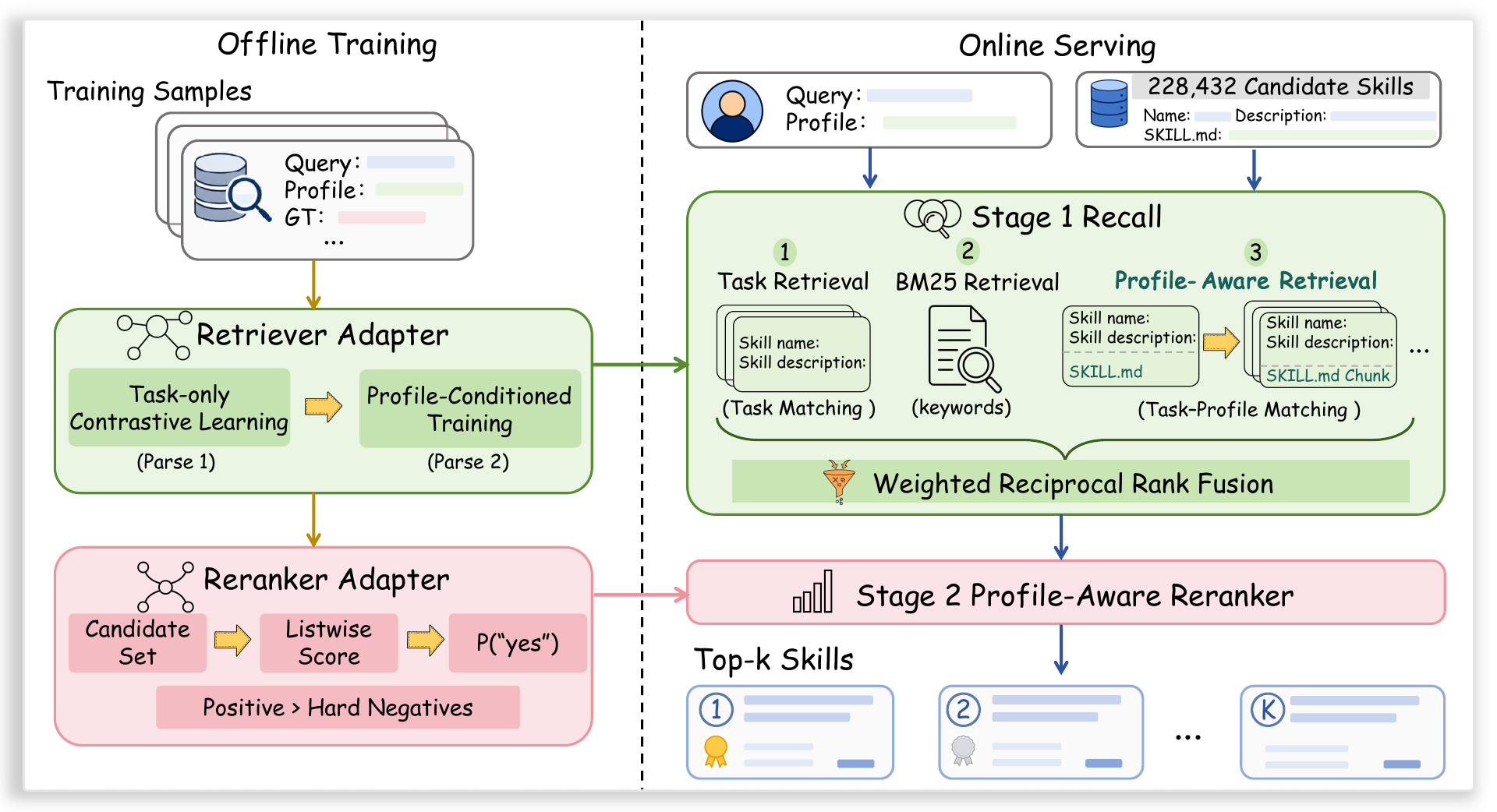}
    \caption{Overview of the proposed personalized skill routing framework. During offline training, user profile-aware supervision is constructed to optimize a dense retriever and a reranker. During online serving, the retriever recalls candidate skills from the repository, and the reranker produces the final personalized ranking.}
    \label{fig:method}
\end{figure}

SkillFeed is designed around three requirements that distinguish profile-conditioned routing from conventional task-skill retrieval. First, a router must retain broad functional coverage over a large repository: profile information should refine task relevance rather than replace it. Second, the decisive profile constraint often separates skills that are already semantically similar, so the model needs supervision in which a fixed task maps to different skills under different profiles. Third, these constraints frequently occur in the procedural body of a skill rather than in its short description. As illustrated in Figure~\ref{fig:method}, SkillFeed addresses these requirements jointly: it constructs counterfactual profile supervision, learns task alignment before profile-sensitive discrimination, and retrieves body-level evidence before making a final profile-conditioned ranking. Thus, the components are not interchangeable retrieval add-ons; each resolves a distinct failure mode of task-only routing.


\subsection{Learning Profile-Conditioned Routing}

\paragraph{Counterfactual Supervision.}

Existing skill retrieval datasets such as SkillRet~\cite{cho2026skillretlargescalebenchmarkskill}, SkillsBench~\cite{li2026skillsbenchbenchmarkingagentskills}, SWE-Skills~\cite{han2026sweskillsbenchagentskillsactually}, and SRA~\cite{su2026skillretrievalaugmentationagentic} provide task--skill pairs, but cannot specify when a profile should overturn an otherwise plausible task-only match. We therefore construct training supervision around task-fixed counterfactual pairs (illustrated in Figure~\ref{fig:benchmark}), separately from the held-out benchmark. Starting from a target skill $S^{+}$, we generate a task $q$ and a compatible profile $p$, then mine task-relevant skills that conflict with $p$ through retrieval and verification as hard negatives set $S^{-}$. For one such skill $\tilde{S}^{+} \in S^{-}$ , we generate a counterfactual profile $\tilde{p}$ under which it becomes the preferred skill for the same $q$ and subsequently mine a corresponding set of hard negatives  $\tilde{S}^{-}$. The resulting paired tuples $(p, q, S^{+}, S^{-})$ and $(\tilde{p}, q, \tilde{S}^{+}, \tilde{S}^{-})$ therefore differ only in the routing-relevant profile but require different rankings. They directly supervise the desired reversal: profile information must discriminate between functionally plausible skills, rather than merely accompany task wording. We remove likely valid alternatives, near duplicates, and invalid skill files to limit false negatives; the resulting corpus contains 2,712 task-centric and 7,459 profile-conditioned instances, with examples in Appendix Table~\ref{tab:training_examples_persona}.

\paragraph{Two-Phase Retriever Training.}
We adapt Qwen3-Embedding-0.6B and Qwen3-Reranker-0.6B~\cite{qwen3embedding} to complementary stages of routing. The retriever operates over the full repository, where a missed functionally relevant skill cannot be recovered, whereas the reranker receives only task-relevant alternatives and can focus on their profile-conditioned differences. Pretrained embeddings capture general semantic similarity rather than skill-specific capability matching. We consequently train the retriever progressively. Phase~1 uses task-only queries and hard negatives to establish routing-specific task--skill alignment. Phase~2 adds profile constraints and the counterfactual pairs above, making the profile a discriminative signal among functionally plausible skills rather than a shortcut for weak task matching. Both phases use InfoNCE~\cite{oord2019representationlearningcontrastivepredictive} over the reference skill, curated hard negatives, and in-batch negatives.

\paragraph{Profile-Aware Reranker Training.}
Unlike the retriever, the reranker is trained directly on profile-conditioned candidate groups. Each group contains one target and task-relevant but profile-conflicting hard negatives; candidates include their names, descriptions, and procedural bodies. We optimize a listwise objective~\cite{10.1145/1390156.1390306} using the model's \textit{yes} logit as the relevance score. This objective reflects the final routing decision: the suitable skill must outrank near misses that satisfy the task in isolation but violate a profile constraint. Complete objectives are provided in Appendix~\ref{app:training_objectives}.

\subsection{Profile-Aware Skill Routing}
\label{subsection:two_stage_inference}

Task relevance alone does not establish profile compatibility. Although two skills may expose the same high-level capability, the conditions that make one appropriate for a particular user often differ significantly. For example, required inputs, supported platforms, output formats, dependencies, and exceptions are often stated only in their procedural bodies. SkillFeed therefore performs \emph{profile-aware skill-body retrieval}: it retrieves localized, profile-relevant conditions from skill body for a task--profile query and converts the retrieved chunks into a skill-level compatibility signal. The signal complements, rather than replaces, task relevance: it identifies which task-relevant skills state execution conditions that fit the requesting profile.

\paragraph{Structured Skill-Body Representation.}
Skill descriptions are concise summaries of core functionality~\cite{openai2026skills}, whereas profile-relevant conditions are distributed across sections such as \textit{When to use}, \textit{Inputs}, \textit{Examples}, \textit{Constraints}, and \textit{Dependencies}. Encoding an entire skill as one document would mix these conditions with unrelated capabilities and dilute a decisive local match. We instead split each file into Markdown-structured chunks, following passage-level retrieval~\citep{lewis2020rag}, and encode them independently. Each chunk retains its skill name, description, section path, and content; long sections are divided into overlapping passages. This representation turns an implicit document-level judgment into retrievable evidence: a profile constraint can match the precise passage that supports or rules out a skill.


\paragraph{Evidence Aggregation.}
For a task--profile query, profile-aware skill-body retrieval returns matching chunks, which we aggregate into a skill-level compatibility score.
For each candidate skill $S_i$, let $\mathcal{C}_{S_i}$ denote the set of
retrieved chunks belonging to $S_i$, and let $h_c$ denote the dense similarity
score of chunk $c$.
Let $\operatorname{TopM}(\mathcal{C}_{S_i})$ denote the set of the $M$
highest-scoring chunks in $\mathcal{C}_{S_i}$. The aggregated score is
computed as 
\begin{equation}
\begin{aligned}
\operatorname{score}(S_i)
={}&
\alpha
\max_{c\in\mathcal{C}_{S_i}} h_c
+
\frac{\beta}{M}
\sum_{c\in\operatorname{TopM}(\mathcal{C}_{S_i})} h_c \\
&+
\gamma
\log\!\left(1+\left|\mathcal{C}_{S_i}\right|\right),
\end{aligned}
\label{eq:chunk_to_skill_aggregation}
\end{equation}
where $\alpha$, $\beta$, and $\gamma$ are non-negative aggregation weights, and $M$ is the number of highest-scoring chunks included in the mean term. The maximum term preserves a single decisive condition, the top-$M$ mean distinguishes repeated evidence from an isolated noisy passage, and the logarithmic count only mildly rewards multiple supporting chunks. This converts local passages into a compatibility signal without mechanically favoring longer skill files.

\paragraph{Integrating Evidence into Routing.}
Profile-aware skill-body retrieval is combined with one profile signal and two task signals: profile dense over profile, task dense retrieval over skill, and BM25 ~\citep{robertson2009bm25} over the same text. The latter preserves exact lexical matches, such as tool names, platforms, and file formats that dense representations may underweight. We combine the three ranked lists with weighted reciprocal rank fusion~\citep{cormack2009rrf}, which avoids calibrating heterogeneous raw scores. The reranker then receives the task, profile, and complete skill information to resolve any remaining conflicts. Thus, task-level retrieval establishes functional plausibility, while profile-aware skill-body retrieval exposes the local conditions needed to choose among plausible skills.

\begin{table*}[!t]
\centering
\resizebox{\linewidth}{!}{%
\begin{tabular}{l c c c c c}
\toprule
\textbf{Method}
& \textbf{Input}
& \textbf{All Hit@1}
& \textbf{Sensitive Hit@1}
& \textbf{Insensitive Hit@1}
& \textbf{CF-Switch Acc.} \\
\midrule

Qwen3-Emb-0.6B + Qwen3-Reranker-0.6B
& Task + Profile
& 0.520
& 0.426
& 0.611
& 0.143 \\

SkillFeed Emb Phase 1
& Task
& 0.483
& 0.340
& 0.623
& 0.052 \\

SkillFeed Emb Phase 2
& Task + Profile
& 0.617
& 0.512
& 0.719
& 0.286 \\

SkillFeed Emb Phase 2 + SkillFeed Reranker
& Task
& 0.644
& 0.444
& 0.838
& 0.026 \\

SkillFeed Emb Phase 2 + SkillFeed Reranker
& Task + Profile
& \textbf{0.733}
& \textbf{0.630}
& \textbf{0.832}
& \textbf{0.377} \\

\bottomrule
\end{tabular}%
}
\caption{Effect of profile conditioning at the retrieval and reranking stages. ``Task'' denotes task-only conditioning, whereas ``Task + Profile'' denotes
profile-aware conditioning. All metrics are computed on 329 test samples,
162 profile-sensitive
samples and 167
profile-insensitive queries.
CF-Switch Acc. denotes Hit@1 on 77 profile-counterfactual samples.}
\label{tab:personalization_matter}
\end{table*}

\begin{table*}[!h]
\centering
\small
\setlength{\tabcolsep}{5pt}
\renewcommand{\arraystretch}{1.08}

\begin{tabular}{@{}l l c c c c c c@{}}
\toprule
\textbf{Embedding Model}
& \textbf{Reranker Model}
& \textbf{Hit@1}
& \textbf{Hit@3}
& \textbf{Hit@5}
& \textbf{Hit@10}
& \textbf{MRR@20}
& \textbf{nDCG@20} \\
\midrule

\multicolumn{8}{@{}l}{\textit{(1) Embedding Only (No Reranker)}} \\
\addlinespace[2pt]

BM25 Only
& ---
& 0.410 & 0.538 & 0.578 & 0.611 & 0.485 & 0.528 \\

E5-large-v2
& ---
& 0.322 & 0.474 & 0.508 & 0.565 & 0.410 & 0.459 \\

BGE-large-en-v1.5
& ---
& 0.386 & 0.492 & 0.556 & 0.626 & 0.461 & 0.509 \\

Qwen3-Emb-0.6B
& ---
& 0.416 & 0.559 & 0.644 & 0.717 & 0.514 & 0.575 \\

SkillFeed Emb Phase 1 (Task-only FT)
& ---
& 0.483 & 0.638 & 0.699 & 0.787 & 0.583 & 0.642 \\

Qwen3-Emb-8B
& ---
& 0.471 & 0.574 & 0.634 & 0.724 & 0.574 & 0.634 \\

SkillFeed Emb Phase 2 (User profile FT)
& ---
& \textbf{0.617}
& \textbf{0.754}
& \textbf{0.796}
& \underline{0.833}
& \textbf{0.697}
& \textbf{0.741} \\

BM25 + Task Dense + Profile Dense
& ---
& 0.492 & 0.641 & 0.705 & 0.815 & 0.597 & 0.663 \\

BM25 + Task Dense + Profile Chunk Dense
& ---
& \underline{0.517}
& \underline{0.705}
& \underline{0.775}
& \textbf{0.860}
& \underline{0.632}
& \underline{0.698} \\

\addlinespace[3pt]
\midrule
\multicolumn{8}{@{}l}{\textit{(2) Embedding + Reranker}} \\
\addlinespace[2pt]

SkillFeed Emb Phase 2
& Qwen3-Reranker-0.6B
& 0.559
& 0.757
& 0.818
& \underline{0.882}
& 0.668
& 0.727 \\

SkillFeed Emb Phase 2
& BGE-reranker-v2-m3
& 0.304
& 0.495
& 0.596
& 0.702
& 0.434
& 0.525 \\

SkillFeed Emb Phase 2
& Qwen3-Reranker-8B
& \underline{0.632}
& \underline{0.772}
& \underline{0.821}
& 0.881
& \underline{0.717}
& \underline{0.763} \\

SkillFeed Emb Phase 2
& LLM (mimo-v2.5-pro)
& 0.590
& 0.726
& 0.769
& 0.824
& 0.672
& 0.715 \\

Qwen3-Emb-8B
& Qwen3-Reranker-8B
& 0.602
& 0.723
& 0.787
& 0.836
& 0.681
& 0.724 \\

SkillRouter Emb 0.6B
& SkillRouter Reranker 0.6B
& 0.614
& 0.726
& 0.751
& 0.772
& 0.675
& 0.703 \\

SkillFeed Emb Phase 2
& SkillFeed Reranker
& \textbf{0.733}
& \textbf{0.818}
& \textbf{0.863}
& \textbf{0.891}
& \textbf{0.788}
& \textbf{0.819} \\

\addlinespace[3pt]
\midrule
\multicolumn{8}{@{}l}{\textit{(3) Full Pipeline (BM25 + Dense + Reranker)}} \\
\addlinespace[2pt]

BM25 + Task Dense + Profile Dense
& SkillFeed Reranker
& \textbf{0.751}
& \underline{0.836}
& \underline{0.869}
& \underline{0.900}
& \textbf{0.803}
& \underline{0.831} \\

BM25 + Task Dense + Profile Chunk Dense
& SkillFeed Reranker
& \underline{0.739}
& \textbf{0.851}
& \textbf{0.888}
& \textbf{0.918}
& \textbf{0.803}
& \textbf{0.838} \\

\bottomrule
\end{tabular}%
\caption[Main results on SkillFeed-Bench.]{
Main results on SkillFeed-Bench.
We organize results by pipeline stage:
(1) Embedding Only,
(2) Embedding + Reranker, and
(3) Full Pipeline.
Bold indicates the best result, and
\underline{underlining} indicates the second-best result.
Task/Profile Dense use SkillFeed Emb Phase 2 for respective inputs; Profile Chunk Dense chunks input.
}
\label{tab:main_results}
\end{table*}

\section{Experiment}

We evaluate three questions. First, does profile conditioning improve routing precisely when a profile changes the suitable skill? Second, how does SkillFeed compare with general-purpose, skill-specific, and larger pretrained retrieval and reranking models? Third, do its retrieval, reranking, and profile-aware skill-body components make complementary contributions? 
Unless otherwise stated, all results are evaluated on SkillFeed-Bench.


\subsection{Experimental Setup}
SkillFeed is initialized from Qwen3-Embedding-0.6B for retrieval and Qwen3-Reranker-0.6B for reranking. We evaluate both a skill-level profile-conditioned dense channel and the proposed profile-aware skill-body retrieval channel; each can be fused with task-oriented dense retrieval and BM25 before reranking. Unless otherwise stated, the system reranks the top 50 fused candidates. Complete training, preprocessing, and inference settings are provided in Appendix~\ref{app:implementation_details}.

\paragraph{Evaluation Protocol.}

We report Hit@K for $K\in\{1,3,5,10\}$, MRR@20, and nDCG@20. Hit@K measures whether the annotated skill is surfaced within the top $K$ results, whereas MRR@20 and nDCG@20 emphasize its early-rank position. To evaluate profile sensitivity, we report results both on the full benchmark and on 162 profile-sensitive queries with  user context constraints. We also refer to Hit@1 on its 77 profile-counterfactual subset as counterfactual-switch accuracy (CF-Switch Acc.).

\paragraph{Baselines.}

We compare SkillFeed with retrieval and reranking baselines corresponding to the two stages of our pipeline.

\textit{Retrievers.}
We include BM25 \cite{robertson2009bm25} as a sparse lexical baseline. E5-large-v2~\cite{wang2022e5} and BGE-large-en-v1.5~\cite{bge_embedding} represent general-purpose dense retrievers. We further evaluate the dense retriever from SkillRouter~\cite{zheng2026skillrouter}, which is specifically trained for task-to-skill routing. Qwen3-Embedding-8B~\cite{qwen3embedding} is included without task-specific adaptation to examine whether model scale alone can replace routing-specific supervision. Finally, we report our Phase~1 retriever to isolate the effect of profile-conditioned adaptation.

\textit{Rerankers.}
We compare our reranker against the pretrained Qwen3-Reranker-0.6B~\cite{qwen3embedding}, BGE-reranker-v2-m3~\cite{chen2024bge}, and the pretrained Qwen3-Reranker-8B~\cite{qwen3embedding}. We additionally use MiMo-V2.5-Pro~\cite{mimo2026v25pro} as LLM reranker, prompting it to assess each candidate with respect to both the task requirements and the profile constraints.

All retrievers search the same 228K-skill repository. For reranker comparisons, every model reranks the same fixed candidate lists, so differences can be attributed to candidate discrimination rather than to recall coverage.

\subsection{Main Results}
\begin{table*}[!h]
\small
\centering
\begin{tabular}{l c c c c}
\toprule
\textbf{Pipeline Configuration} & \textbf{Hit@1} & \textbf{Hit@5} & \textbf{Hit@10} & \textbf{MRR@20} \\
\midrule
\multicolumn{5}{l}{\textit{Single-Stage Retrieval}} \\
BM25 Only & 0.410 & 0.578 & 0.611 & 0.485 \\
SkillFeed Emb Phase 2 & 0.617 & 0.796 & 0.833 & 0.697 \\
\midrule
\multicolumn{5}{l}{\textit{Two-Stage: Dense + Reranker}} \\
SkillFeed Emb Phase 2 + Qwen3-Reranker-0.6B & 0.559 & 0.818 & 0.882 & 0.668 \\
SkillFeed Emb Phase 2 + SkillFeed Reranker & 0.733 & 0.863 & 0.891 & 0.788 \\
\midrule
\multicolumn{5}{l}{\textit{Full Pipeline: BM25 + Dense + Reranker}} \\
Full Pipeline (topk=50) & \textbf{0.751} & \textbf{0.869} & \textbf{0.900} & \textbf{0.803} \\
Full Pipeline (topk=100) & 0.739 & 0.866 & 0.903 & 0.796 \\
\midrule
\multicolumn{5}{l}{\textit{Ablation: Pipeline Components}} \\
Full Pipeline w/o BM25 & 0.723 & 0.878 & 0.909 & 0.788 \\
Full Pipeline w/o Profile Dense & 0.726 & 0.854 & 0.891 & 0.785 \\
\bottomrule
\end{tabular}%
\caption[Ablation on retrieval pipeline.]{Ablation on retrieval pipeline. The full pipeline defaults topk=50. ``w/o'' indicates specific components are removed.}
\label{tab:ablation_pipeline}
\end{table*}
%

\subsubsection{Necessity of Personalized Skill Routing.}
Table~\ref{tab:personalization_matter} presents a controlled comparison between task-only and profile-conditioned routing configurations. The pretrained Qwen baseline and the task-only Phase~1 retriever achieve only 0.143 and 0.052 Hit@1 on profile-counterfactual queries, despite obtaining 0.611 and 0.623 on profile-insensitive queries. This substantial gap demonstrates that task semantics alone are insufficient for personalized skill routing and confirms the need for profile-counterfactual evaluation. Incorporating the constructed profile-aware supervision in Phase~2 improves counterfactual Hit@1 from 0.052 to 0.286. Moreover, adding user profile conditioning to the reranker further increases counterfactual Hit@1 from 0.026 to 0.377, while insensitive-query performance changes only marginally from 0.838 to 0.832. These findings demonstrate that personalized skill routing is necessary under profile shifts: the counterfactual construction in SkillFeed-Bench exposes skill changes that task-level evaluation overlooks, while fine-tuning the retriever and reranker on profile-aware data with counterfactual supervision enables them to effectively recognize and resolve these changes.
\subsubsection{Overall Comparison.}

Table~\ref{tab:main_results} compares SkillFeed with representative baselines at three increasingly complete settings: embedding-only retrieval, retrieval followed by reranking, and the full hybrid system. This organization separates the contribution of routing-specific representation learning, profile-aware candidate discrimination, and complementary candidate-generation signals.

\textit{Profile-Aware Embedding and Reranking.}
SkillFeed Emb Phase~2 achieves the best embedding-only performance, reaching 0.617 Hit@1, compared with 0.483 for the task-only Phase~1 retriever and 0.471 for the unfine-tuned Qwen3-Embedding-8B model. Other general-purpose baselines perform substantially worse, with BM25, E5-large-v2, and BGE-large-en-v1.5 achieving only 0.410, 0.322, and 0.386 Hit@1, respectively. The adapted 0.6B model also outperforms the pretrained 8B model, indicating that domain-specific supervision can offset model-size differences. For reranking, the SkillFeed reranker raises Hit@1 to 0.733 with the Phase~2 retriever, outperforming the Qwen3-Reranker-8B and SkillRouter pipelines by 10.1 and 11.9 percentage points, respectively. These results show that profile-aware supervision improves both candidate retrieval and fine-grained reranking, providing gains beyond task-only training, general-purpose models, and increased model scale.

\textit{Multi-Source Recall and Profile Chunk Retrieval.}
The full pipeline achieves the best Hit@1 of 0.751 and MRR@20 of 0.803 by combining profile-aware dense retrieval, BM25, and SkillFeed reranking. The Profile  chunk-based variant slightly lowers Hit@1 to 0.739 but improves Hit@10 to 0.918 and nDCG@20 to 0.838. This trade-off reflects the division of labor between the two stages: retrieval prioritizes placing the ground-truth skill within the top-$K$ candidate set, whereas reranking determines its final position. Overall, the results validate the complementary benefits of profile-aware training, multi-source recall, and skill-aware chunk retrieval.

\subsection{Further Analysis}

\paragraph{Retrieval Pipeline Ablation.}
Table~\ref{tab:ablation_pipeline} shows that replacing the pretrained Qwen3 reranker with the SkillFeed reranker improves Hit@1 from 0.559 to 0.733, demonstrating the importance of routing-specific reranking. The full pipeline further reaches 0.751 Hit@1 and 0.803 MRR@20, confirming that multi-source recall and profile-aware reranking provide complementary benefits. Removing profile dense retrieval reduces both Hit@1 and MRR@20, validating its role in retrieving persona-relevant candidates. Removing BM25 lowers early-rank precision but slightly improves deeper recall, indicating that lexical retrieval provides complementary but rank-sensitive evidence. The top-50 setting offers the best balance between early precision and deeper coverage.

\paragraph{Performance Across User Profile Groups.}
Appendix Table~\ref{tab:persona_analysis} analyzes the model across country, language, and domain groups. While SkillFeed achieves 0.733 overall Hit@1, its lower performance on specialized technical and scientific queries indicates that fine-grained domain and user profile constraints remain challenging. At the same time, the results demonstrate that the proposed framework generalizes across heterogeneous user profiles rather than being limited to a single user profile category.

\paragraph{Statistical Significance.}
Tables~\ref{tab:bootstrap_ci} and~\ref{tab:permutation_test} in the appendix report bootstrap confidence intervals and paired permutation tests. The full system achieves Hit@1 of 0.751 with a 95\% confidence interval of $[0.706, 0.794]$ and improves on the pretrained Qwen baseline by 0.231 ($p<0.0001$). The gains from the SkillFeed reranker over embedding-only retrieval (+0.116, $p=0.001$) and from including the profile in reranking (+0.089, $p<0.01$) are also significant. These tests support the reliability of the central comparisons reported above.

\section{Conclusion}

This work reframes skill routing in LLM agents as a profile-conditioned relevance problem. We show that task semantics alone cannot reliably identify
the correct skill when user constraints induce different routing decisions: on the profile-sensitive queries in SkillFeed-Bench, task-only retrieval achieves only 0.426  Hit@1, while personalized retrievers raise it to 0.630. SkillFeed, our two-stage framework combining multi-source candidate recall with profile-conditioned reranking, achieves 0.751 Hit@1 overall, with the largest gains concentrated on profile-sensitive instances. These results establish user profile information as an essential signal for personalized skill routing in agentic systems. 
Future work will explore dynamically extracting and updating user constraints from multi-turn conversational contexts~\cite{liu2026diagnostic}.

\clearpage

\bibliography{aaai2027}

\lstnewenvironment{appendixcode}
  {\lstset{basicstyle=\footnotesize\ttfamily,
    numbers=none,
    xleftmargin=0pt,
    aboveskip=2pt,
    belowskip=2pt,
    showstringspaces=false,
    tabsize=2,
    breaklines=true,
    breakatwhitespace=false,
    columns=fullflexible,
    keepspaces=true}}
  {}

\clearpage
\onecolumn
\appendix

\section{benchmark }

 .

\subsection{Test Sample Format}

Each test sample in SkillFeed-Bench is a JSON line in \texttt{test.jsonl}. The top-level structure is shown below:
\begin{appendixcode}
{
  "task": {
    "name": "...",
    "description": "...",
    "core_requirements": ["...", "..."]
  },
  "user_context_constraints": {
    "country_region": "...",
    "language": "...",
    "preferred_platforms": ["..."],
    "payment_methods": ["..."],
    "other_constraints": ["..."]
  },
  "ground_truth_skill": {...},
}
\end{appendixcode}
The \texttt{task} object describes the user's natural-language request; \texttt{user\_context\_constraints} captures the profile attributes (country, language, platform, payment, domain constraints); \texttt{ground\_truth\_skill} identifies the single correct skill from the 228K candidate pool. The 329 test samples include 162 profile-sensitive
queries (77 profile-counterfactual augmentation) and 167
profile-insensitive queries.

Each retained original and counterfactual instance was independently reviewed by two annotators with prior experience in AI agents, software tools, and programming-related tasks. We provided annotation guidelines specifying three criteria: (1) whether the skill is capable of completing the task, (2) whether user profile information changes the preferred skill, and (3) whether the counterfactual profile preserves the original task while introducing a meaningful preference shift. Disagreements were resolved by an expert reviewer using the same criteria, and ambiguous cases without clear consensus were discarded. This verification process ensures that benchmark labels reflect profile-conditioned skill suitability rather than superficial lexical overlap.
\begin{figure}[htbp]
    \centering
    \includegraphics[width=0.8\columnwidth]{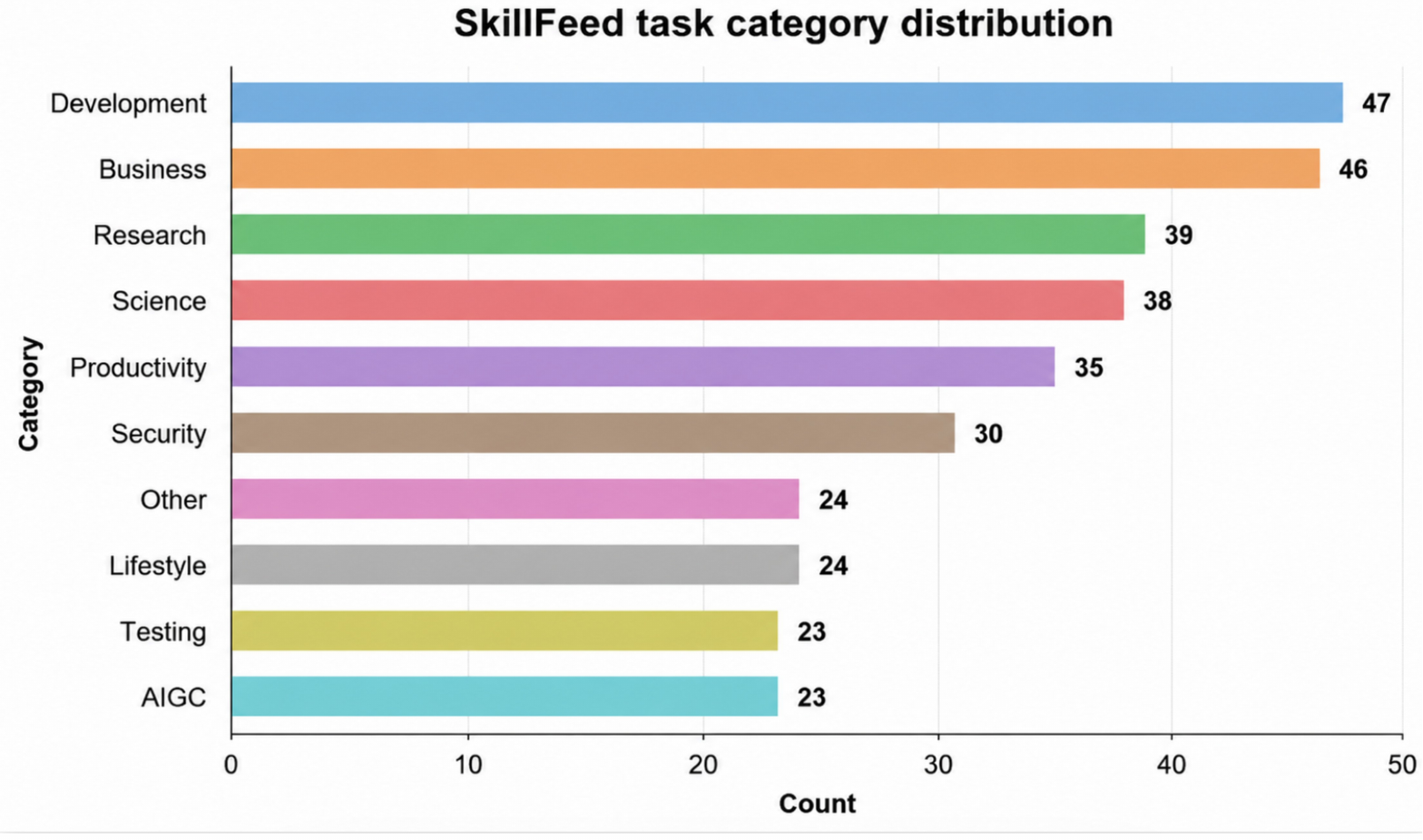}
    \caption{SkillFeed-Bench consists of tasks spanning 10 categories.}
    \label{fig:category_distribution}
\end{figure}

Table~\ref{tab:dataset_stats} summarizes the key statistics. The test set consists of 329 samples drawn from a candidate pool of 228,432 skills. Each sample is annotated with 4 hard negative skills that are semantically similar but profile-incompatible. The average task description is 61.8 words, and the average skill body excerpt is 16,730 characters. To prevent trivial lexical matching, we enforce a token overlap threshold of $\leq 0.50$ between the task description and the ground-truth skill body, with an average overlap of 0.29. The annotator confidence averages 0.92, indicating high annotation quality.

\begin{table}[h]
\centering
\caption{Key statistics of SkillFeed-Bench.}
\label{tab:dataset_stats}
\begin{tabular}{l r}
\toprule
\textbf{Statistic} & \textbf{Value} \\
\midrule
Total skill candidates & 228,432 \\
Test samples & 329 \\
\quad -- Profile-sensitive & 162 \\
\quad \qquad -- Profile counterfactual & 77 \\
\quad \qquad -- Standard (with constraints) & 85 \\
\quad -- Profile-insensitive & 167 \\
Unique skill categories & 10 \\
Avg.\ task description length (words) & 61.8 \\
Avg.\ skill body excerpt length (chars) & 16,730 \\
Token overlap (task--skill body) & 0.29 (max 0.50) \\
Annotator confidence (avg.) & 0.92 \\
\bottomrule
\end{tabular}
\end{table}

The test set covers 10 skill categories spanning a diverse range of domains. As shown in Table~\ref{fig:category_distribution}, the two largest categories are Development (47 samples, 14.3\%) and Business (46 samples, 14.0\%), followed by Research (39, 11.9\%) and Science (38, 11.6\%). The remaining six categories---Productivity, Security, Other, Lifestyle, Testing, and AIGC---each contribute 23--35 samples (7.0\%--10.6\%), ensuring a balanced coverage across application domains.

\section{Training and Pipeline}
\subsection{Learning Profile-Conditioned Routing}

\paragraph{User Profile-Aware Supervision Construction.}

Existing skill retrieval datasets such as SkillRet~\cite{cho2026skillretlargescalebenchmarkskill}, SkillsBench~\cite{li2026skillsbenchbenchmarkingagentskills}, SWE-Skills~\cite{han2026sweskillsbenchagentskillsactually}, and SRA~\cite{su2026skillretrievalaugmentationagentic} are valuable for general skill retrieval, but they are insufficient for modeling personalized skill routing. To address this limitation, as illustrated in the right-hand flow of Figure~\ref{fig:benchmark}, we construct a training dataset that incorporates both profile-based and counterfactual supervision.

Starting from a target skill $S^{+}$, an LLM generates a realistic task query $q$ and a compatible profile $p$ based on the skill description and procedural content. The profile captures routing-relevant constraints, such as language, region, preferred platform, and payment method. We then construct a hard-negative set $\mathcal{S}^{-}$ by combining semantic retrieval with LLM-based verification, prioritizing skills that are relevant to the task but incompatible with the user profile. This process yields task-centric supervision tuples $(p,q,S^{+},\mathcal{S}^{-})$.

To strengthen user profile-conditioned supervision, we further perform target-first counterfactual augmentation. Specifically, a hard negative $\tilde{S}^{+}\in\mathcal{S}^{-}$ is selected as the counterfactual target, and an LLM generates a new profile $\tilde{p}$ under which $\tilde{S}^{+}$ is preferred for the same query, yielding $(\tilde{p},q,\tilde{S}^{+},\tilde{\mathcal{S}}^{-})$. These paired instances provide direct supervision for learning how profile changes affect skill preference while the task remains unchanged. Finally, we filter likely valid alternatives, near-duplicate skills, and candidates without valid skill content to reduce false negatives. The resulting dataset contains 2,712 task-centric and 7,459 user profile-conditioned instances, with representative examples shown in Appendix Table~\ref{tab:training_examples_persona}.

\begin{table*}[t]
\centering
\small
\begin{tabular}{p{0.12\linewidth} p{0.40\linewidth} p{0.40\linewidth}}
\toprule
& \textbf{Phase 1: Task-Only Training Data} & \textbf{Phase 2: Profile-Conditioned Training Data} \\
\midrule

\textbf{Task} &
\multicolumn{2}{p{0.82\linewidth}}{
I want to plan a trip to Shanghai including attractions and a simple schedule.
}
\\

\midrule

\textbf{User Profile} &
(Not included in query)
&
Occupation: Student \\
&
&
Budget: Low budget \\
&
&
Transportation: Subway preferred
\\

\midrule

\textbf{Query Format} &
\begin{tabular}[t]{@{}p{\linewidth}@{}}
Task name: Shanghai Travel Planning\\
Task description: Plan a trip to Shanghai including attractions and a simple itinerary.\\
\end{tabular}
&
\begin{tabular}[t]{@{}p{\linewidth}@{}}
Task name: Shanghai Travel Planning\\
Task description: Plan a trip to Shanghai including attractions and a simple itinerary.\\
User/context constraints:\\
occupation: Student\\
budget: Low budget\\
transportation: Subway preferred
\end{tabular}
\\

\midrule

\textbf{Ground Truth} &
\texttt{city-travel-guide} 
&
\texttt{budget-travel-planner} 
\\

\midrule

\textbf{Why GT is correct} &
The query only specifies the travel planning task, so a general-purpose travel planning skill is sufficient.
&
The user profile indicates limited budget and preference for public transportation, making a budget-oriented planner the most suitable skill.
\\

\midrule

\textbf{Hard Negative} &
\texttt{attraction-recommender} 
&
\texttt{luxury-itinerary-planner} 

\\

\midrule

\textbf{Why HN is wrong} &
The skill focuses only on attraction recommendation and cannot generate a complete travel itinerary.
&
Assumes premium travel preferences and taxi transportation, which conflict with the user profile.
\\

\bottomrule
\end{tabular}
\caption{Training data examples for two-stage skill retrieval. Phase~1 learns general task-to-skill matching using only the task description, while Phase~2 incorporates user profile constraints to enable personalized skill selection.}
\label{tab:training_examples_persona}
\end{table*}

\paragraph{Two-Phase Retriever Training.}
Using the constructed supervision dataset, we adapt Qwen3-Embedding-0.6B and Qwen3-Reranker-0.6B~\cite{qwen3embedding} to their complementary roles in personalized skill routing. The dense retriever must preserve broad task--skill relevance while incorporating user profile information into candidate recall, whereas the reranker focuses on fine-grained user profile-conditioned discrimination among semantically relevant candidates.


We optimize the retriever in two phases. Although the pretrained embedding model captures general semantic similarity, it is not specialized for skill routing, which requires matching task requirements to the capabilities and applicability conditions encoded in skill documents. Phase~1 therefore uses task-only queries and hard negatives to learn routing-specific task--skill alignment. Phase 2 appends profile constraints to the query and introduces paired instances in which the same task is associated with different reference skills. This schedule first adapts the model to skill routing and then learns user profile-conditioned discrimination.

For each training instance, let $u_i$ denote the query representation,
$S_i^{+}$ the reference skill, and $\mathcal{N}_i$ the set of curated
and in-batch negative skills. In Phase~1, the query contains only the task,
i.e., $u_i=q_i$. In Phase~2, the user profile is appended to the task,
i.e., $u_i=[q_i;p_i]$.

We optimize the dense retriever with the InfoNCE objective \cite{oord2019representationlearningcontrastivepredictive}:
\begin{equation}
\mathcal{L}_{\mathrm{ret}}
=
-\frac{1}{B}
\sum_{i=1}^{B}
\log
\frac{
\exp\!\left(
\operatorname{sim}
\left(
E_{\theta}(u_i),
E_{\theta}(S_i^{+})
\right)/\tau_{\mathrm{ret}}
\right)
}{
\sum\limits_{S\in\{S_i^{+}\}\cup\mathcal{N}_i}
\exp\!\left(
\operatorname{sim}
\left(
E_{\theta}(u_i),
E_{\theta}(S)
\right)/\tau_{\mathrm{ret}}
\right)
},
\label{eq:app_retriever_loss}
\end{equation}
where $E_{\theta}$ is the dense encoder,
$\operatorname{sim}(\cdot,\cdot)$ denotes cosine similarity,
$B$ is the batch size, and $\tau_{\mathrm{ret}}$ is the temperature.
The objective encourages the reference skill to be ranked above skills that
are task-relevant but incompatible with the user profile.

\paragraph{Persona-Aware Reranker Training.}
Unlike the dense retriever, the reranker is trained directly with user profile-aware supervision because it operates on a small candidate set whose members are already relevant to the task. Its primary role is therefore to determine which candidate best satisfies the user profile-specific constraints. For each query containing both task and profile information, we construct a ranking group consisting of one positive skill and multiple curated hard negatives. Each candidate is represented by its name, description, and procedural body.

The reranker operates on a candidate group containing one reference skill
$S_i^{+}$ and a set of hard negatives $\mathcal{N}_i$. Given query $u_i$
and candidate skill $S$, Qwen3-Reranker produces a relevance score
$r_{\phi}(u_i,S)$ from the logit assigned to the token \textit{yes}.
We optimize the model using listwise cross-entropy:
\begin{equation}
\mathcal{L}_{\mathrm{rank}}
=
-\frac{1}{B}
\sum_{i=1}^{B}
\log
\frac{
\exp\!\left(
r_{\phi}(u_i,S_i^{+})/\tau_{\mathrm{rank}}
\right)
}{
\sum\limits_{S\in\{S_i^{+}\}\cup\mathcal{N}_i}
\exp\!\left(
r_{\phi}(u_i,S)/\tau_{\mathrm{rank}}
\right)
},
\label{eq:app_reranker_loss}
\end{equation}
where $\tau_{\mathrm{rank}}$ is the ranking temperature. Normalizing scores
within each candidate group directly trains the reference skill to outrank
semantically similar alternatives.

\subsection{Persona-Aware Routing Pipeline}
\label{subsection:two_stage_inference}

At inference time, SkillFeed adopts a two-stage inference pipeline that separates broad candidate coverage from fine-grained candidate discrimination.
The first stage retrieves candidate skills from the 228K-scale repository, and the second stage reranks the retrieved candidates.
This separation allows the recall stage to focus on preserving potentially relevant skills, while the reranker determines which candidate best satisfies both the task requirement and the user profile.

\paragraph{Candidate Recall.}
The recall stage combines three complementary retrieval channels.
The first channel is task-oriented dense retrieval, which encodes the task name and task description as the query and retrieves skills according to their names and descriptions.
The second channel applies BM25 over the same skill-level text, providing a lexical matching signal complementary to dense retrieval~\citep{robertson2009bm25}.
The third channel is persona-conditioned dense retrieval over structured chunks extracted from \texttt{SKILL.md} bodies.
Unlike the first two channels, which rely on skill-level summaries, this channel exposes body-level evidence during recall and is designed to capture fine-grained user profile-related information.

Skill descriptions are usually concise and mainly summarize the core function of a skill \cite{openai2026skills}.
However, whether a skill is suitable for a particular user often depends on details that may appear only in the skill body, such as applicable scenarios, input requirements, output formats, tool dependencies, constraints, and examples.
These details are commonly distributed across sections such as \textit{When to use}, \textit{Inputs}, \textit{Examples}, \textit{Constraints}, and \textit{Dependencies}.
Therefore, relying only on skill names and descriptions may miss candidates whose main function is relevant but whose user profile-specific applicability is expressed in the body text.

\paragraph{Skill-to-Chunk Representation.}
A straightforward alternative is to embed the entire \texttt{SKILL.md} file as a single document.
However, this may produce overly coarse representations, since long-document embeddings tend to mix multiple capabilities, conditions, and constraints.
As a result, localized but important matching signals may be diluted.
Following passage-level retrieval in retrieval-augmented generation~\citep{lewis2020rag}, we instead split each \texttt{SKILL.md} file into Markdown-structured chunks and encode these shorter passages independently.


For each chunk, we store the skill name, description, section path, and content.
Long sections are further divided into overlapping passages, so that local applicability conditions can be recovered during recall.
This chunk-level index enables profile-conditioned retrieval to match user preferences and constraints against specific skill-body evidence, rather than only against high-level skill summaries.

\paragraph{Chunk-to-Skill Aggregation.}
User profile-conditioned retrieval returns chunk-level results, which are then aggregated into skill-level candidates before fusion.
For each candidate skill $S_i$, let $\mathcal{C}_{S_i}$ denote the set of
retrieved chunks belonging to $S_i$, and let $h_c$ denote the dense similarity
score of chunk $c$.
Let $\operatorname{TopM}(\mathcal{C}_{S_i})$ denote the set of the $M$
highest-scoring chunks in $\mathcal{C}_{S_i}$. The aggregated score is
computed as
\begin{equation}
\begin{aligned}
\operatorname{score}(S_i)
={}&
\alpha
\max_{c\in\mathcal{C}_{S_i}} h_c
+
\frac{\beta}{M}
\sum_{c\in\operatorname{TopM}(\mathcal{C}_{S_i})} h_c \\
&+
\gamma
\log\!\left(1+\left|\mathcal{C}_{S_i}\right|\right),
\end{aligned}
\label{eq:chunk_to_skill_aggregation}
\end{equation}
where $\alpha$, $\beta$, and $\gamma$ are non-negative aggregation weights,
and $M$ is the number of highest-scoring chunks included in the mean
term. The maximum term captures the strongest local match between the user profile and a passage in the skill body. 
The top-$M$ mean term measures whether the skill is supported by multiple highly relevant chunks, thereby reducing the influence of an isolated noisy match. 
The logarithmic count term provides a mild reward for skills associated with multiple retrieved chunks.
Overall, this aggregation favors skills with consistent body-level evidence while preventing an excessive preference for skills containing a larger
number of chunks.

\paragraph{Fusion and Profile-Aware Reranking.}
The three recall lists are combined using weighted reciprocal rank fusion~\citep{cormack2009rrf}.
The fused top candidates are then passed to the trained reranker, which takes the task, the user profile, and the full candidate skill information as input.
In this pipeline, chunk-based retrieval is used to improve user profile-aware candidate coverage during recall.
It does not directly determine the final ranking; instead, it increases the likelihood that skills with relevant body-level evidence are included for subsequent fine-grained reranking.

\section{Additional Experimental Details}
\label{app:experimental_details}

\subsection{Dataset Structure and Input Formatting}
\label{app:dataset_format}

\paragraph{Underlying Sample Schema.}
Each source instance contains a task, an optional user profile, one reference
skill, and a set of hard-negative skills:
\begin{appendixcode}
{
  "task": {
    "name": "...",
    "description": "...",
    "core_requirements": ["...", "..."]
  },
  "user_context_constraints": {
    "country_region": "...",
    "language": "...",
    "preferred_platforms": ["..."],
    "payment_methods": ["..."],
    "other_constraints": ["..."]
  },
  "ground_truth_skill": {...},
  "hard_negative_skills": [{...}, {...}]
}
\end{appendixcode}

Each skill record contains a name, description, and a
\texttt{SKILL.md} body excerpt. The reference skill is treated as the
single annotated positive for training and evaluation.

\paragraph{Phase 1: Task-Only Supervision.}
Phase~1 removes the user profile from the query and uses task-centric
supervision. The query instruction is:
\begin{quote}
\small
\texttt{Represent this user task for retrieving the most appropriate skill.}
\end{quote}
The resulting query is formatted as:
\begin{appendixcode}
Represent this user task for retrieving the most appropriate skill.

Task name: {task_name}

Task description: {task_description}

Core requirements:
- {requirement_1}
- {requirement_2}
\end{appendixcode}
The training record follows the ms-swift embedding schema:
\begin{appendixcode}
{
  "messages": [{"role": "user", "content": "<query>"}],
  "positive_messages": [
    [{"role": "user", "content": "<positive_skill>"}]
  ],
  "negative_messages": [
    [{"role": "user", "content": "<negative_skill_1>"}],
    [{"role": "user", "content": "<negative_skill_2>"}]
  ]
}
\end{appendixcode}

Phase~1 contains 2,712 training instances and 356 validation instances.

\paragraph{Phase 2: Profile-Conditioned Supervision.}
Phase~2 retains the same task fields and appends user-context constraints:
\begin{appendixcode}
Represent this user task for retrieving the most appropriate skill.

Task name: {task_name}

Task description: {task_description}

Core requirements:
- {requirement_1}
- {requirement_2}

User/context constraints:
country_region: {region}
language: {language}
preferred_platforms: {platforms}
payment_methods: {payment_methods}
other_constraints: {constraints}
\end{appendixcode}

Its negatives combine curated hard negatives, LLM-guided
profile-conflicting negatives, and counterfactual positives from other
profile variants of the same task. The Phase~2 corpus contains 7,459
training instances and 608 validation instances, including full-body and
profile-counterfactual supervision.

\paragraph{Skill Document Format.}
Positive and negative skills are serialized with the same field order:
\begin{appendixcode}
Skill name: {skill_name}

Skill description: {skill_description}

Skill body:
{skill_body_excerpt}
\end{appendixcode}
The body excerpt exposes procedural details and applicability constraints
that are not always present in the short description.

\paragraph{Reranker Records.}
Reranker training uses the task--profile query together with a candidate
group containing one positive and multiple negatives. The generative
reranker record is:
\begin{appendixcode}
{
  "system":
    "Judge whether the Document meets the requirements based
     on the Query and the Instruct provided. Note that the
     answer can only be \"yes\" or \"no\".",
  "messages": [{"role": "user", "content": "<query>"}],
  "positive_messages": [
    [{"role": "assistant", "content": "<positive_skill>"}]
  ],
  "negative_messages": [
    [{"role": "assistant", "content": "<negative_skill_1>"}],
    [{"role": "assistant", "content": "<negative_skill_2>"}]
  ]
}
\end{appendixcode}

The document role is \texttt{user} for embedding records and
\texttt{assistant} for generative reranker records. During reranker
training and inference, the system instruction, query template, skill field
names, and constraint serialization are kept consistent to avoid formatting
artifacts.

\subsection{Training Objectives}
\label{app:training_objectives}

\subsection{Implementation Details}
\label{app:implementation_details}

The SkillFeed dense retriever is initialized from
Qwen3-Embedding-0.6B, and the SkillFeed reranker is initialized from
Qwen3-Reranker-0.6B. Both models are fine-tuned on 2$\times$A100 GPUs using
AdamW with a learning rate of $1\times10^{-5}$, an effective batch size of 16,
a maximum sequence length of 2048, and three training epochs. The retriever is
trained in two phases: 2,712 task-centric instances are used for task-only
alignment, followed by 7,459 user profile-conditioned instances for
profile-aware adaptation. Lexical-overlap filtering uses a threshold of 0.50.

For \texttt{SKILL.md} chunk retrieval, Markdown sections are divided into
passages with a target length of 200--500 tokens, a maximum length of 800
tokens, and a small overlap between adjacent passages. Chunk-to-skill
aggregation uses $\alpha=0.70$, $\beta=0.25$, $\gamma=0.05$, and $M=3$.
Weighted reciprocal rank fusion assigns weights of 0.40 to task dense
retrieval, 0.40 to profile-conditioned dense retrieval over
\texttt{SKILL.md} chunks, and 0.20 to BM25. Unless otherwise stated, the full
pipeline reranks the top 50 fused candidates; the top-100 setting is evaluated
only in the ablation study.

\begin{table}[!htbp]
\centering
\caption[Per-Profile-group analysis]{Per-Profile-group analysis on SkillFeed Emb Phase 2 + SkillFeed Reranker with Profile. We evaluate performance across different persona groups: country/region, language, domain, and persona sensitivity. All 329 test samples are included.}
\label{tab:persona_analysis}
\begin{tabular}{l c c c c c}
\toprule
\textbf{Group} & \textbf{N} & \textbf{Hit@1} & \textbf{Hit@5} & \textbf{MRR@20} & \textbf{nDCG@20} \\
\midrule
\multicolumn{6}{l}{\textit{By Country/Region}} \\
Global/Unknown & 257 & 0.735 & 0.879 & 0.795 & 0.827 \\
United States & 46 & 0.717 & 0.804 & 0.760 & 0.785 \\
China & 14 & 0.571 & 0.714 & 0.648 & 0.696 \\
Other Regions & 12 & 0.917 & 0.917 & 0.917 & 0.917 \\
\midrule
\multicolumn{6}{l}{\textit{By Language}} \\
English & 204 & 0.672 & 0.804 & 0.735 & 0.770 \\
Unknown & 100 & 0.870 & 0.980 & 0.907 & 0.927 \\
Programming & 11 & 0.545 & 0.818 & 0.632 & 0.677 \\
Other Languages & 14 & 0.786 & 0.929 & 0.845 & 0.866 \\
\midrule
\multicolumn{6}{l}{\textit{By Domain}} \\
General & 248 & 0.790 & 0.895 & 0.836 & 0.860 \\
Tech & 44 & 0.614 & 0.818 & 0.692 & 0.738 \\
Science & 15 & 0.533 & 0.800 & 0.663 & 0.727 \\
Other Domains & 22 & 0.455 & 0.682 & 0.530 & 0.577 \\
\midrule
\multicolumn{6}{l}{\textit{Overall}} \\
\textbf{Overall} & \textbf{329} & \textbf{0.733} & \textbf{0.863} & \textbf{0.788} & \textbf{0.819} \\
\bottomrule
\end{tabular}%
\end{table}


\begin{table}[!htbp]
\centering
\caption{Bootstrap 95\% confidence intervals for key configurations.}
\label{tab:bootstrap_ci}
\begin{tabular}{l c c c c}
\toprule
\textbf{Configuration} & \textbf{Hit@1} & \textbf{95\% CI} & \textbf{MRR@20} & \textbf{95\% CI} \\
\midrule
Full Pipeline & 0.751 & [0.706, 0.794] & 0.803 & [0.760, 0.846] \\
SkillFeed Emb Phase 2 + SkillFeed Reranker w Profile & 0.733 & [0.684, 0.778] & 0.788 & [0.744, 0.832] \\
SkillFeed Emb Phase 2 + SkillFeed Reranker w/o Profile & 0.644 & [0.596, 0.696] & 0.712 & [0.663, 0.761] \\
SkillFeed Emb Phase 2  & 0.617 & [0.568, 0.669] & 0.697 & [0.647, 0.746] \\
SkillFeed Emb Phase 1 & 0.483 & [0.429, 0.535] & 0.583 & [0.530, 0.636] \\
\bottomrule
\end{tabular}%
\end{table}

\begin{table}[!htbp]
\centering
\caption[Permutation test results for key comparisons.]{
Permutation test results for key comparisons.
Statistical significance: $^{*}p<0.05$, $^{**}p<0.01$, and $^{***}p<0.001$.
}
\label{tab:permutation_test}
\begin{tabular}{l c c c}
\toprule
\textbf{Comparison}
& \textbf{Hit@1 $\Delta$}
& \textbf{$p$-value}
& \textbf{Sig.} \\
\midrule

Full Pipeline vs Qwen3-Emb-0.6B + Qwen3-Reranker-0.6B
& +0.231 & $<0.0001$ & $^{***}$ \\

Embedding Only vs Embedding + SkillFeed Reranker
& +0.116 & 0.001 & $^{***}$ \\

Reranker without vs with Profile
& +0.089 & $<0.01$ & $^{**}$ \\

\bottomrule
\end{tabular}%
\end{table}

\end{document}